\documentclass[11pt, a4paper]{article}
\usepackage[bottom, multiple]{footmisc}
\usepackage[document]{ragged2e}
\usepackage[margin=1 in]{geometry}
\usepackage[titletoc]{appendix}
\usepackage[fleqn,tbtags]{mathtools}
\usepackage{amssymb}
\usepackage{amsmath}
\usepackage{amsfonts}
\usepackage{bm}
\usepackage{bbm}

\usepackage{enumitem}
\usepackage{float}
\usepackage{tikz}
\usepackage{csquotes}
\usepackage[justification=centering, skip=2pt]{caption}

\usepackage{multirow}
\usepackage{xcolor,colortbl}
\usepackage{verbatim}
\usepackage{nicefrac}
\usepackage{longtable}
\usepackage{multicol}
\usepackage{xcolor}
\usepackage{sectsty}
\usepackage[backend=biber, style=numeric, url=false]{biblatex}
\nocite{*}

\usepackage[indent]{parskip}
\usepackage{setspace}
\usepackage{hyperref}
\hypersetup{
    colorlinks=true, 
    linktoc=all, 
    citecolor=blue,
    urlcolor=red,
    linkcolor=blue,  
}

\renewenvironment{abstract}
               {\list{}{\rightmargin\leftmargin}%
                \item[]\small\textbf{\normalsize Abstract:}}
               {\endlist}

\begin{document}
\title{\textbf{Is Shapley Explanation for a model unique?}}
\author{Harsh Kumar\thanks{Email: harshk7797@gmail.com, harsh.kumar2@in.ey.com} \quad \quad Jithu Chandran\thanks{Email: jithuc47@gmail.com, jithu.chandran@in.ey.com}}
\date{\vspace{-2em}}
\maketitle
\justifying
\begin{abstract}
Shapley value has recently become a popular way to explain  the predictions of complex and simple machine learning models. This paper is discusses the factors that influence Shapley value. In particular, we explore the relationship between the distribution of a feature and its Shapley value. We extend our analysis by discussing the difference that arises in Shapley explanation for different predicted outcomes from the same model. Our assessment is that Shapley value for particular feature not only depends on its expected mean but on other moments as well such as variance and there are disagreements for baseline prediction, disagreements for signs and most important feature for different outcomes such as probability, log odds, and binary decision generated using same linear probability model (logit/probit). These disagreements not only stay for local explainability but also affect the global feature importance. We conclude that there is no unique Shapley explanation for a given model. It varies with model outcome (Probability/Log-odds/binary decision such as accept vs reject) and hence model application.

\end{abstract}

\normalsize
\large

\section{Introduction}
Machine learning models are visible today in a wide variety of fields. From the field of medicine to data mining, speech recognition to human computer interaction, financial investment to risk management, these complex and hard-to-interpret models outperform traditional models by utilizing intricate algorithms for predictions. Explaining these complex models is crucial not only to understand model predictions but also facilitate in explaining which features are prime contributors to the prediction. Recently, Shapley value has become a popular way to explain the predictions of machine learning models due to a series of desirable theoretical properties which distinguish it from other explanation methods such as LIME \cite{lundberg2017unified}\cite{ribeiro2016should}. Game theoretic Shapley value describes a way to distribute total gains of the cooperative game among players in such a way that it satisfies few desirable notions of fairness \cite{shapley195317}\cite{young1985monotonic}. Shapley value for a player is equal to average of marginal contribution of player over all possible ways in which collation can be formed. In Machine learning models, Shapley value of a feature equals the average marginal contribution of a feature over all possible permutations of features.
 
Shapley value imparts a lens to understand convoluted machine learning models by providing marginal contribution of each feature. This value not only depends on the feature values and prediction function but also on the distribution of data. Although we get machine learning model explanation using Shapley value but what are the factors that influence Shapley value? How does distribution of a feature influence its Shapley value? How does Shapley explanation vary for different predicted outcomes from the same model (e.g. logit model has three output log-odds, probability and binary decision such as accept vs reject)? To the best of our knowledge, these questions are still unexplored in literature.
 
In general these questions are difficult to answer as Shapley value does not have a closed form solution and numerical estimates for Shapley explanation are computationally expensive. Thus, in this paper we tackle these questions for linear probability model (logit/probit). A linear probability model such as logit regression has three outcome types- log-odds, probability, and binary decision. These outcomes can be used for different business purposes. For example, a bank can use probability outcome to estimate expected default rate or use binary decision outcome of the same model to accept or reject a credit application. It is important to know whether Shapley explanation for all the different outcomes of a model are aligned. If not, how are they different? The rest of this paper is organized as follows. Section \ref{sec:logit_probit_model} covers description of the Logit and Probit model with different outcomes that can be generated using the same model. In Section \ref{sec:shapley_value}, we have provided a closed form solution for the Shapley value and have discussed factors influencing these values. Section \ref{sec:disagreements} gives us a snapshot of disagreements that arise between the Shapley value for different outcomes. Section \ref{sec:diagreements_variance} contains information about how disagreements are influenced by the variance of a distribution. Section \ref{sec:global_feature_importance} conveys information about global importance of a feature for different outcomes. Finally, Section \ref{sec:conclusion} holds some concluding remarks. 

\section{Logit and Probit Model} \label{sec:logit_probit_model}
Logit and Probit models are commonly used in industry and academics to predict binary dependent variable $y$ using explanatory variables $X$. Given the coefficient of explanatory variables $\beta$, the probability of $y = 1$ is given by following expression
\[p = Pr(y=1 | X) = G(X\beta)\]
where, $G$ is standard logistic function in case of logit regression and standard normal function in case of probit regression. 

Let, $\eta$ equals $X\beta$. We can interpret $\eta$ as log odds\footnote{log odds equals to $\log\left(\dfrac{p}{1-p}\right)$} for logit model and distance from mean in standard deviation unit for probit model. In general, we use probability models for classification or binary decision making by setting a threshold either on $p$ or $\eta$. Without loss of generality, assume $\eta^*$ is the threshold for binary decision making. Using $\eta^*$, we can get equivalent threshold in a probability 
\[p^* = G(\eta^*)\]

We can interpret the logit/probit model outcome in three different ways:
\begin{itemize}
    \item \textbf{Log-Odds or Standard Deviation Unit} ($\eta$): Outcome $\eta$ can be interpreted as log-odds and distance from mean in standard deviation unit for the logit and probit models, respectively. Since  $\eta$ is linear in $X$ it is easy to interpret and Shapley value has a closed form solution based on $\beta$ and  mean of a feature, irrespective of distribution of explanatory variables for a given sample.
    \item \textbf{Probability} ($p$):  
    The prediction can also be interpreted in terms of probability which is extensively used in literature and industry. Probability prediction outcome captures changes in $\eta$ more effectively than log odds outcome because all changes in $\eta$ will not uniformly translate to probability.
    \item \textbf{Binary Decision Making} ($\mathbbm{1}\left(\eta \ge \eta^*\right)$ or $\mathbbm{1}\left(p \ge p^*\right)$): This interpretation of logit/probit  outcome is directly related to binary decision making scenario. For example: Acceptance or rejection of credit card application.
\end{itemize}


\section{Shapley Value} \label{sec:shapley_value}
Shapley value method was initially introduced in game theory to describe a way to distribute the total gains of the cooperative game among players which satisfies few desirable notions of fairness. Shapley value for a player is equal to the average of marginal contribution over all possible ways in which collation can be formed. Strumbel \& Kononenko \cite{strumbelj2010efficient}\cite{strumbelj2014explaining} and Lundberg \& Lee \cite{lundberg2017unified} described a way to use Shapley value for machine learning model explanation.

Let $f(x^*)$ be  a machine learning model, where $x^*$ represents the data with $M = \{1, 2, ..., m\}$ features. For a sample $x$, Shapley values for the features $i \in M$ denoted by $\phi_i$ are additive, i.e.,
\[f(x) - \phi_0 = \sum_{i \in M} \phi_i\]
where, $\phi_0 $ equals the expected prediction value i.e., $\mathbb{E}[f(X)]$. Thus, Shapley value explains the difference between prediction of the sample and the global average. Shapley explanation for a feature $i$ is estimated in the following way. 
\begin{align*}
    \phi_i =\sum_{S \subseteq M \setminus \{i\}} \frac{|S|! (m-|S|-1)!}{m!}(v(S\cup\{i\})-v(S))
\end{align*}
where, $v(S)$ equals to the expected predicted value using $S$ features i.e., $ \mathbb{E}[f(X)|X_S = x_s]$.

In the case of two features $X_1$ and $X_2$, Shapley value expressions are:
\begin{align*}
    \phi_0 &= v(\{\}) = \mathbb{E}[f(X)]\\
    \phi_1 &= \dfrac{1}{2}\left\{\left[v(\{X_1\}) - v(\{\})\right] + \left[v(\{X_1, X_2\}) - v(\{X_2\})\right]\right\}\\
    \phi_2 &= \dfrac{1}{2}\left\{\left[v(\{X_2\}) - v(\{\})\right] + \left[v(\{X_1, X_2\}) - v(\{X_1\})\right]\right\}
\end{align*}

\subsection{Shapley Value for Linear Probability Model}
For simplicity, we assume two independent\footnote{Under the independence assumption both interventions and conditional expectation are same, this isolates the difference arising from different notation of expectation} and normally distributed  features $X_1$ and $X_2$ with mean $\mu_i$ and variance $\sigma^2_i$. We denote value function and Shapley value with $v$ and $\phi$ respectively where superscripts $\eta$, $p$, and $d$ denote log-odds (or standard deviation unit), probability and binary decision outcomes, respectively and subscript 0 denotes baseline expectation (expected output); 1 and 2 stand for features.

Since both explanatory variables $X_1$ and $X_2$ are normal independent variables, the conditional distribution of $\eta$ or $X\beta$ is 
\begin{align*}
    X\beta &\sim \mathcal{N}(\beta_0 + \beta_1\mu_1 + \beta_2\mu_2, \ \beta^2_1\sigma^2_1 + \beta^2_2\sigma^2_2) \\
    X\beta|(X_1 = x_1)&\sim \mathcal{N}(\beta_0 + \beta_1x_1 + \beta_2\mu_2, \ \beta^2_2\sigma^2_2)\\ 
    X\beta|(X_2 = x_2) &\sim \mathcal{N}(\beta_0 + \beta_1\mu_1 + \beta_2x_2, \ \beta^2_1\sigma^2_1) \\
    X\beta|(X_1 = x_1, X_2 = x_2) &= \beta_0 + \beta_1x_1 + \beta_2x_2
\end{align*}
Above expressions illustrate that conditioning on features, changes the mean of $\eta$ and reduces variance (reduction in uncertainty over $\eta$).

In Appendix \ref{sec:appendix_shapley_value}, we have derived expressions for the value function and Shapley value. For logit regression, value function for the probability outcome requires estimation of logistic function (also known as sigmoid function, we will denote this by $S$) over a normal distribution, i.e., $E[S(Z)]$ where $Z$ is normally distributed but it does not have a closed form solution. Thus, we use standard normal distribution approximation for the logistic/sigmoid function given by the expression below \cite{tocher}\cite{dombi2018approximations}.
\[S(x) \approx \Phi\left(\dfrac{x}{\sqrt{\nicefrac{8}{\pi}}}\right) \]

\begin{figure}[H]
    \centering
    \caption{Outcome Function (For $\eta^* = 0$)}
    \includegraphics[scale=0.45, trim={0 0 0 1.3cm}, clip]{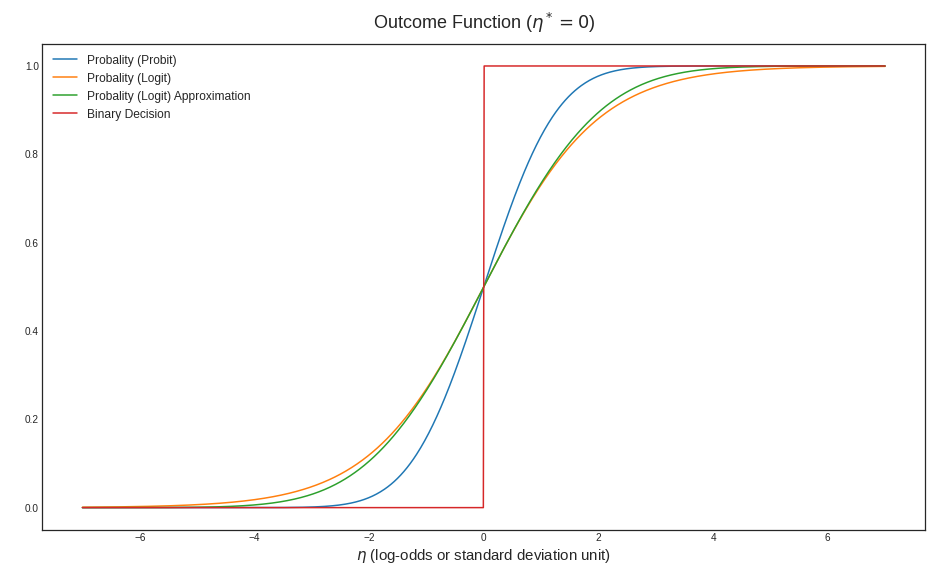}  
    \label{fig:outcome_function}
\end{figure}

Figure \ref{fig:outcome_function} depicts binary and probability outcomes as a function of $\eta$. It also portrays that the standard normal approximation of logistic function is close to actual logistic function.

\begin{figure}[H]
    \centering
    \caption{Feature 1 Shapley value level curves for $\eta$}
    \includegraphics[scale=0.52, trim={0 0 2.5cm 1.3cm}, clip]{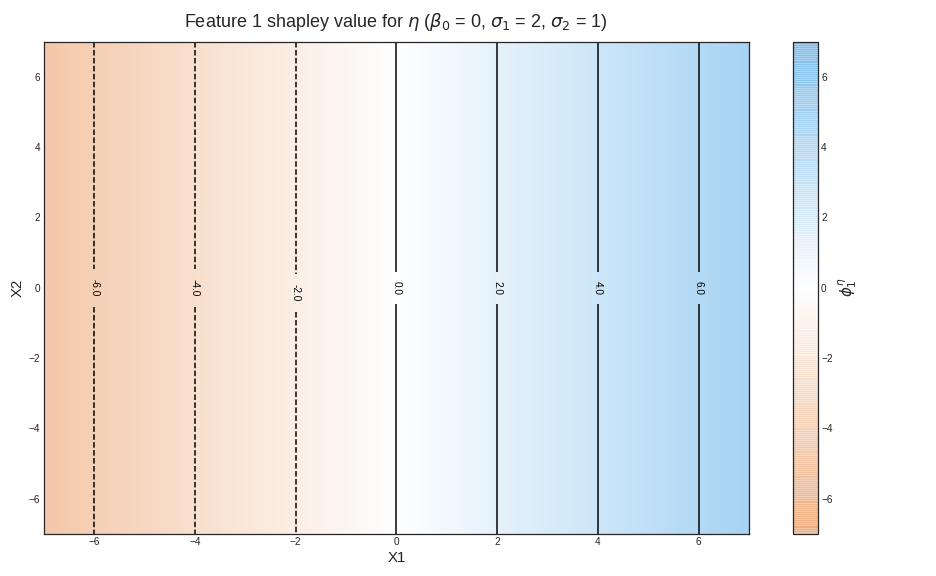}  
    \label{fig:phi1_eta_beta0_0}
\end{figure}

When both features are relevant (slope coefficients are non-zero), we can transform/normalize the data in such a way that features have  0 mean and their regression coefficients are 1 with non-negative intercept. Thus, without loss of generality, we assume $\mu_1 = \mu_2 = 0$, $\beta_1 = \beta_2 = 1$ and $\beta_0 \ge 0$. We also assume $\sigma_1 > \sigma_2$. Under these conditions and $\eta^* = 0$ (for comparison), Shapley value expressions derived in Appendix \ref{sec:appendix_shapley_value} are shown below-
\vspace{1cm}\\
{\Large \textbf{Shapley value for} $\eta$:}
\begin{align*}
    \phi^\eta_0 &= \beta_0\\
    \phi^\eta_1 &= x_1 \tag*{(S1)}\label{eq:S1} \\
    \phi^\eta_2 &= x_2
\end{align*}
Baseline Shapley value for $\eta$ (log-odds or standard deviation units)  is equal to intercept whereas Shapley value for a particular feature is linear in its value and does not depend on the value of the other features. Since, we have linearly transformed\footnote{$\phi_i = \beta_i(x_i - \mu_i)$ for $i \in \{1, 2\}$} the variable to have mean 0 and slope coefficient 1, Shapley value for a feature has same sign as slope coefficient whenever feature value is above its mean. Shapley value for a feature increases in magnitude as we move away from its mean. Figure \ref{fig:phi1_eta_beta0_0} illustrates Shapley value for feature 1 as a function of $x_1$ and $x_2$ for log odd output. We can observe that the level curves for $\phi_1$  are linear in $x_1$ and do not depend on $x_2$.

{\Large \textbf{Shapley value for Probability} ($p$):}
\begin{align*}
    \phi^p_0 &= \Phi\left(\dfrac{\beta_0}{\sqrt{\lambda + \sigma^2_1 + \sigma^2_2}}\right) \tag*{(S2)}\label{eq:S2}\\
    \phi^p_1 &= \dfrac{1}{2}\left\{\left[\Phi\left(\dfrac{\beta_0 + x_1 }{\sqrt{\lambda + \sigma^2_2}}\right) - \Phi\left(\dfrac{\beta_0}{\sqrt{\lambda + \sigma^2_1 + \sigma^2_2}}\right)\right] + \left[\Phi\left(\dfrac{\beta_0 + x_1 + x_2}{\sqrt{\lambda}}\right) - \Phi\left(\dfrac{\beta_0 + x_2}{\sqrt{\lambda + \sigma^2_1}}\right)\right]\right\} \\
    \phi^p_2 &= \dfrac{1}{2}\left\{\left[\Phi\left(\dfrac{\beta_0 + x_2 }{\sqrt{\lambda + \sigma^2_1}}\right) - \Phi\left(\dfrac{\beta_0}{\sqrt{\lambda + \sigma^2_1 + \sigma^2_2}}\right)\right] + \left[\Phi\left(\dfrac{\beta_0 + x_1 + x_2}{\sqrt{\lambda}}\right) - \Phi\left(\dfrac{\beta_0 + x_1}{\sqrt{\lambda + \sigma^2_2}}\right)\right]\right\}
\end{align*}

In \ref{eq:S2}, $\lambda$  equal to 1 in case of Probit model ($G = \Phi$) and $\lambda = \nicefrac{8}{\pi} \approx 2.5465$  in case of the Logit model ($G = S$). It is not reasonable to directly compare logit and probit model with same parameters $\beta$, as from the above approximation we can infer that the parameters of logit model will be approximately scaled by $\sqrt{\nicefrac{8}{\pi}} \approx 1.5958$ compared to probit model parameters. Thus, we will avoid comparing logit and probit model results.

\begin{figure}[H]
    \centering
    \caption{Feature 1 Shapley value for probability(probit) ($\beta_0$ = 0, $\sigma_1$ = 2, $\sigma_2$ = 1)}
    \includegraphics[scale=0.52, trim={0 0 2.5cm 1.3cm}, clip]{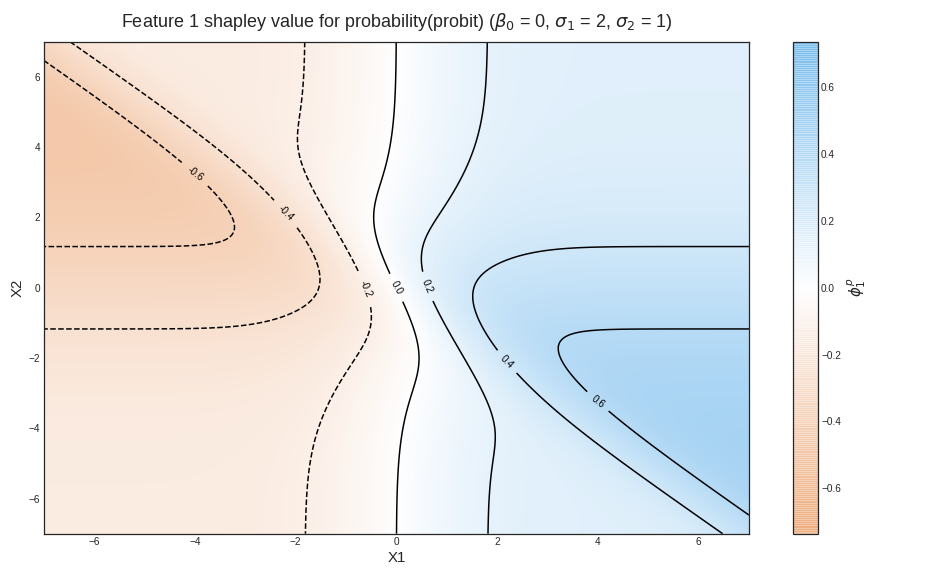}  
    \label{fig:phi1_p_beta0_0}
\end{figure}

Above equations demonstrate that  the Shapley value of a feature for probability outcome depends on all feature values and variance.  Intuitively variance is important for a value function which is conditional expectation of probability over $\eta$. The spread of observation around the mean depends on variance and absolute changes in probability due to increase or decrease in $\eta$ by the same constant are not equal. Hence, positive and negative spread from mean are not equally influencing value function implying  value function (in turn Shapley value) depends on variance. For example, when variance is close to 0, the value function equals to $G(\mathbb{E}[\eta])$ but for high variance and positive $\beta_0$, positive spread from mean will have less influence compared to negative spread implying a reduction in value function. Figure \ref{fig:phi1_p_beta0_0} illustrates, Shapley value for feature 1 ($\phi_1^p$) as a function of $x_1$ and $x_2$ in the case of probit model. It indicates that the level curve for $\phi_1^p$ are non linear in both $x_1$ and $x_2$ and Shapley value for feature 1 will always increase with increase in feature 1 value but it can go in either way if we increase feature 2 value. This can also be inferred from the above expression.


{\Large \textbf{Shapley value for Binary Decision} ($\mathbbm{1}\left(\eta \ge \eta^*\right)$ or $\mathbbm{1}\left(p \ge p^*\right)$):}
\begin{align*}
    \phi^d_0 &= \Phi\left(\dfrac{\beta_0}{\sqrt{\sigma^2_1 +\sigma^2_2}}\right) \tag*{(S3)}\label{eq:S3}\\
    \phi^d_1 &= \dfrac{1}{2}\left\{\left[\Phi\left(\dfrac{\beta_0 + x_1}{\sigma_2}\right) - \Phi\left(\dfrac{\beta_0}{\sqrt{\sigma^2_1 + \sigma^2_2}}\right)\right] + \left[\mathbbm{1}\left(\beta_0 + x_1 + x_2 \ge 0 \right) - \Phi\left(\dfrac{\beta_0 + x_2}{\sigma_1}\right)\right]\right\} \\
    \phi^d_2 &= \dfrac{1}{2}\left\{\left[\Phi\left(\dfrac{\beta_0 + x_2}{\sigma_1}\right) - \Phi\left(\dfrac{\beta_0}{\sqrt{\sigma^2_1 + \sigma^2_2}}\right)\right] + \left[\mathbbm{1}\left(\beta_0 + x_1 + x_2 \ge 0 \right) - \Phi\left(\dfrac{\beta_0 + x_1}{\sigma_2}\right)\right]\right\}
\end{align*}

\begin{figure}[H]
    \centering
    \caption{Feature 1 Shapley value for binary decision ($\beta_0$ = 0, $\sigma_1$ = 2, $\sigma_2$ = 1)}
    \includegraphics[scale=0.52, trim={0 0 2.5cm 1.3cm}, clip]{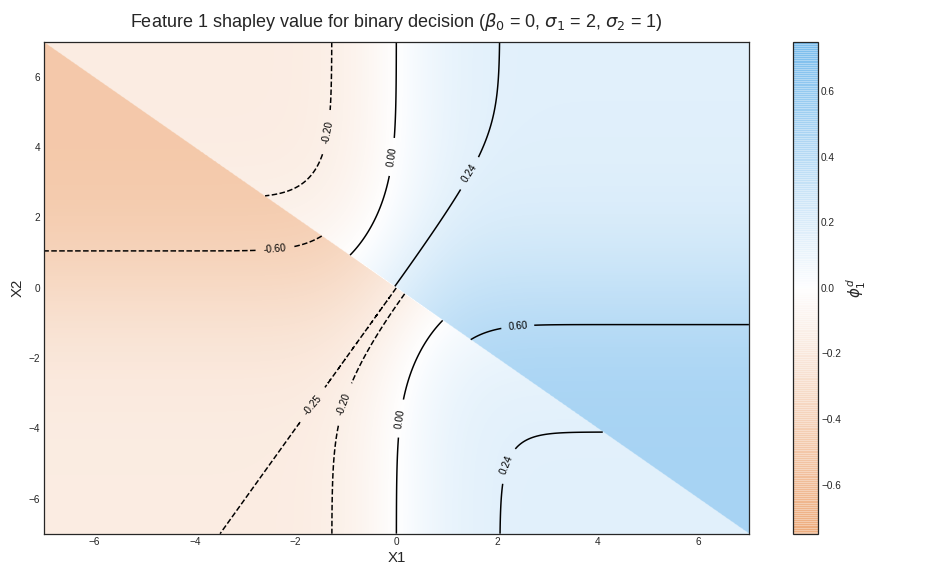}  
    \label{fig:phi1_d_beta0_0}
\end{figure}

Shapley value for binary decision outcome also depends on other feature values and their variance like probability outcome. Shapley value for binary decision is discontinuous at $\beta_0 + x_1 + x_2 = \eta^*$ due to the presence of a binary indicator variable for decision. Thus, Shapley value for positive decision and negative decision can be significantly different even for similar feature values that are close to cut off. In other words, a minor change in sample can lead to a significant difference in Shapley value. Figure \ref{fig:phi1_d_beta0_0}  highlights that the level curves for $\phi_1^d$ are discontinuous at the decision boundary.

\section{Disagreements}\label{sec:disagreements}
In the last section, we observed that Shapley values for different outcomes have different expressions and level curves. For example, level curves for outcome $\eta$ are vertical lines, for probability they are non linear, and for binary decision, they are non-linear and discontinuous. Due to these differences, Shapley value for different outcomes is expected to have few disagreements. In this section we will focus on few major disagreements described below:
\begin{itemize}
    \item \textbf{Disagreement in baseline expectation}: The baseline expectation for an outcome is equal to $\phi_0$. Since the sum of Shapley value for all features is equal to prediction over and above the average value, disagreement in baseline expectation implies different reference points for Shapley value generated using different outcomes, which makes them incomparable with each other. 
    \item \textbf{Disagreement in sign of Shapley value}: Disagreement in sign of a Shapley value using different outcomes indicates a  same feature is positively contributing for one outcome and negatively for the other. For example, feature 1 is positively contributing for probability outcome and negatively contributing for binary decision. Given, both probability and binary decision are monotonic transformation of $\eta$, this disagreement between signs is counter intuitive especially in a case where there is no disagreement in baseline expectation. 
    \item \textbf{Disagreement in most important feature}: Feature with the highest absolute Shapley value can be interpreted as the most important feature and knowing the important feature plays an important role in understanding model prediction. A disagreement about the most important feature for different outcomes from the same model indicates that a feature can be most important for one outcome but not for the other.
\end{itemize}

These disagreements are discussed below in detail with the help of plots to highlight the region of disagreement.
 
\subsection{Disagreement in Baseline Expectation}
The baseline value for Shapley explanation is $\phi_0$ or $v(\{\})$ i.e., unconditional expected value of the outcome. Baseline expectation $\phi_0^p$ and $\phi_0^d$ are measured in probability but $\phi_0^\eta$ is measured in log-odds/standard-deviation units. Thus, it is not wise to directly compare the baseline expectation of three outcomes. Due to this, we will transform $\eta$ using $G$ such that its baseline expectation is comparable with other outcomes, here $G$ is standard normal $\Phi$ in case of probit and logistic function $S$ or $\Phi(\nicefrac{.}{\sqrt{\nicefrac{8}{\pi}}})$ in case of logit regression. If there is no disagreement in baseline expectation, we have
\begin{align*}
    &\mathbb{E}(p) = \mathbb{P}(\eta \ge 0) = G(\mathbb{E}(\eta))\\
    \Longleftrightarrow \ & \phi_0^p = \phi_0^d = \Phi\left(\dfrac{\phi_0^\eta}{\sqrt{\lambda}}\right)\\
    \Longleftrightarrow \ & \Phi\left(\dfrac{\beta_0}{\sqrt{\lambda + \sigma^2_1 + \sigma^2_2}}\right) = \Phi\left(\dfrac{\beta_0}{\sqrt{\sigma^2_1 + \sigma^2_2}}\right) = \Phi\left(\dfrac{\beta_0}{\sqrt{\lambda}}\right)
\end{align*}
where, $\lambda$  equals to 1 in case of Probit model ($G = \Phi$) and $\lambda = \nicefrac{8}{\pi} \approx 2.5465$ for the Logit model ($G = S$).

For $\beta_0 = 0$, the baseline expectation for $\eta$ is 0 and baseline expectation for probability/binary-decision is 0.5 implying all three baseline expectation are aligned, i.e.,
\begin{align*}
    \mathbb{E}(p) = \mathbb{P}(\eta \ge 0) = G(\mathbb{E}(\eta))  = 0.5
\end{align*}

For $\beta_0 > 0$, the baseline expectation $\phi_0$ (or $v(\{\})$) for different outcomes are not aligned. Specifically, for  $\sigma^2_1 + \sigma^2_2 > \lambda $ (other case only affects the right most inequality)
\begin{align*}
    &0.5 < \Phi\left(\dfrac{\beta_0}{\sqrt{\lambda + \sigma^2_1 + \sigma^2_2}}\right) < \Phi\left(\dfrac{\beta_0}{\sqrt{\sigma^2_1 + \sigma^2_2}}\right) <\Phi\left(\dfrac{\beta_0}{\sqrt{\lambda}}\right)\\
    \implies& 0.5 < \mathbb{E}(p) < \mathbb{P}(\eta \ge 0) < \Phi(\mathbb{E}(\eta))
\end{align*}

Intuitively, the difference between the baseline of log-odds/standard-deviation unit ($\eta$) and probability arises because the function $\Phi$ is concave in $\mathbb{R}^+$, convex in $\mathbb{R}^-$ and has point symmetry ($180^\circ$ rotation symmetry). We know for a concave function, the expected value lies below the function  value evaluated at expected value and converse is true in case of Convex. Because we have taken the underlying distribution to be symmetric, when $\beta_0 > 0$ most of the observations lie on the concave region and hence, the expected probability is less than the probability at expected $\eta$ (log-odds or standard deviation unit). When $\beta_0 = 0$, observations are symmetrically distributed in concave and convex region, thus we have expected probability equal to the probability at expected $\eta$ (log-odds or standard deviation unit).

In the real world, we commonly have data with unbalanced classes and observing a class with probability more than 0.95 is not rare which implies high $\mathbb{E}(\eta)$ or $\beta_0 > 0$. Thus, we could easily observe significant difference between $\mathbb{E}(p)$, $\Phi(\mathbb{E}(\eta))$ and $\mathbb{P}(\eta \ge 0)$ this difference increases as probability of common class increases.

\subsection{Disagreement in Sign}
Given, there is no closed form solution for a feature Shapley value equal to zero ($\phi_i = 0$) in the case of probability and binary decision outcomes. First, we will mathematically argue that there exists a point around the mean of the explanatory variable where there is disagreement of sign between Shapley value for $\eta$ and probability/binary decision when $\beta_0 > 0$. Later we will graphically highlight the region even for the case when $\beta_0 = 0$ (no baseline difference).

\textbf{Shapley value for $\eta$}
\begin{align*}
    \phi^\eta_1 = \phi^\eta_2 = 0
\end{align*}

\textbf{Shapley value for $p$}
\begin{align*}
    \phi^p_1 &= \dfrac{1}{2}\left\{\left[\Phi\left(\dfrac{\beta_0}{\sqrt{\lambda +  \sigma^2_2}}\right) - \Phi\left(\dfrac{\beta_0}{\sqrt{\lambda + \sigma^2_1 + \sigma^2_2}}\right)\right] + \left[\Phi\left(\dfrac{\beta_0}{\lambda}\right) - \Phi\left(\dfrac{\beta_0}{\sqrt{\lambda + \sigma^2_1}}\right)\right]\right\} > 0\\
    \phi^p_2 &= \dfrac{1}{2}\left\{\left[\Phi\left(\dfrac{\beta_0}{\sqrt{\lambda +  \sigma^2_1}}\right) - \Phi\left(\dfrac{\beta_0}{\sqrt{\lambda + \sigma^2_1 + \sigma^2_2}}\right)\right] + \left[\Phi\left(\dfrac{\beta_0}{\lambda}\right) - \Phi\left(\dfrac{\beta_0}{\sqrt{\lambda + \sigma^2_2}}\right)\right]\right\} > 0
\end{align*}

\textbf{Shapley value for Binary Decision}
\begin{align*}
    \phi^d_1 &= \dfrac{1}{2}\left\{\left[\Phi\left(\dfrac{\beta_0 }{\sigma_2}\right) - \Phi\left(\dfrac{\beta_0}{\sqrt{\sigma^2_1 + \sigma^2_2}}\right)\right] + \left[1 - \Phi\left(\dfrac{\beta_0}{\sigma_1}\right)\right]\right\} > 0\\
    \phi^d_2 &= \dfrac{1}{2}\left\{\left[\Phi\left(\dfrac{\beta_0 }{\sigma_1}\right) - \Phi\left(\dfrac{\beta_0}{\sqrt{\sigma^2_1 + \sigma^2_2}}\right)\right] + \left[1 - \Phi\left(\dfrac{\beta_0}{\sigma_2}\right)\right]\right\} > 0
\end{align*}

Above expressions illustrate the value of Shapley explanation for both features at mean and  $\beta_0 > 0$. Shapley values for probability and binary decision for both features are positive whereas Shapley explanation for $\eta$ (log-odds/standard-deviation unit) is 0 for both features. Intuitively, this happens because there are two effects- 
\begin{itemize}
    \item \textbf{Change in Expected $\eta$ (Log-Odds or Standard deviation unit)}: This effect accounts for the contribution of variables that arise due to change in expected value of $\eta$. This effect is present in Shapley explanation for all outcomes. But at the mean of a feature this effect is 0 as conditioning on a feature does not change the mean of conditional distribution. This is the only effect present in the Shapley value for $\eta$. Thus, we have 0 Shapley value for both features for outcome $\eta$ at mean. 
    \item \textbf{Reduction in Uncertainty of  $\eta$ (Log-Odds or Standard deviation unit)}: From earlier discussion/value function expression, we noticed that the value function for probability and binary decision also depends on variance. Reduction in the variance positively affects the value function (for $\beta_0 > 0$). When we condition on a variable, it reduces the variance and positively affects the value function. Because of this reason we have a positive Shapley value for probability and binary decision.
\end{itemize}

Since, Shapley explanation for a feature is monotonic in  its value and continuous around mean for $\beta_0 > 0$\footnote{Shapley value for binary decision is continuous around mean for $\beta_0 > 0$ because the decision does not change around mean which is the only reason of discontinuity}. There exists a point around the mean such that the Shapley explanation in log-odds has opposite sign compared to Shapley value for probability and binary decision because the latter is strictly positive. Intuitively, at this point reduction in uncertainty effect dominates the change in expected $\eta$ effect.

\begin{figure}[H]
    \centering
    \caption{Level Curve for zero Shapley value ($\beta_0$ = 0, $\sigma_1$ = 2, $\sigma_2$ = 1)} 
    \includegraphics[scale=0.45, trim={0 0 0 1.3cm}, clip]{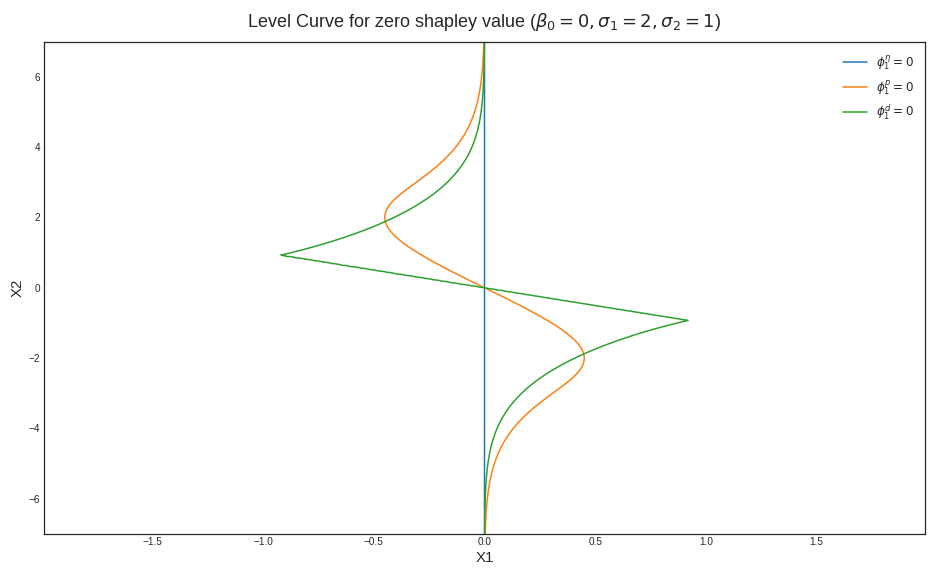}
    \label{fig:level_curve_zero_shap_beta0_0}
\end{figure}

\begin{figure}[H]
    \centering
    \caption{Level Curve for zero Shapley value ($\beta_0$ = 1, $\sigma_1$ = 2, $\sigma_2$ = 1)}
    \includegraphics[scale=0.45, trim={0 0 0 1.3cm}, clip]{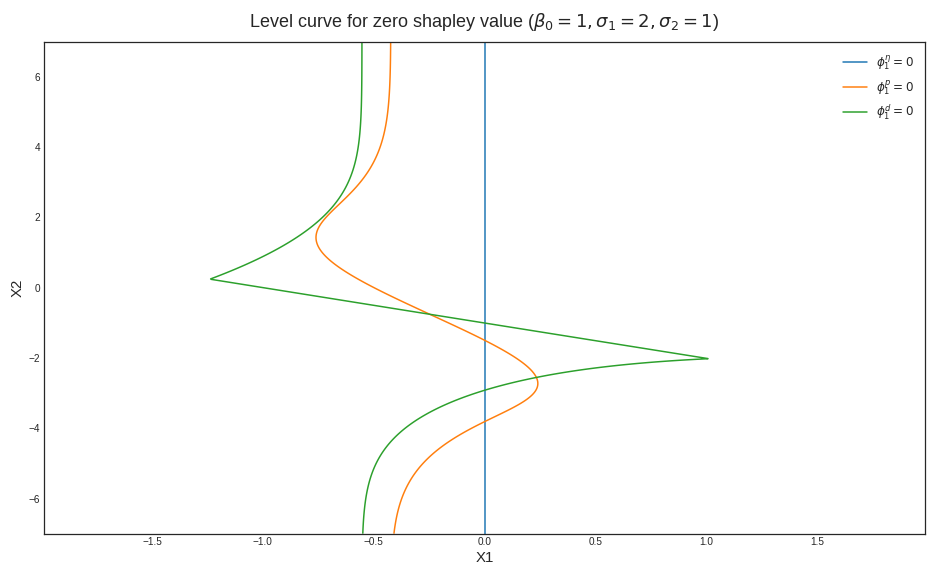}
    \label{fig:level_curve_zero_shap_beta0_1}
\end{figure}

Figure \ref{fig:level_curve_zero_shap_beta0_0} and \ref{fig:level_curve_zero_shap_beta0_1} illustrate the level curves for zero Shapley value for different outcomes for feature 1\footnote{Shapley value for binary decision is not exactly 0 for points overlapping with decision boundary ($\beta_0 + x_1 + x_2 = \eta^*$). But it separates the region of positive and negative Shapley value}. Shapley value for feature 1  is positive for all points to  the right of the level curve and negative to the left. It  also illustrates the region where  feature 1 Shapley explanation is positive for one outcome and negative for other even when there is no baseline difference. For example any point in a region between right of a green line and left of an orange has positive Shapley value for probability but negative for binary decision. Visually, this region for sign disagreement is higher for $\beta_0$ equal to 1.

\subsection{Disagreement in Most Important Feature}
We refer to the feature with highest absolute Shapley value as the most important feature. In order to avoid the explanation difference due to baseline expectation difference, we start with $\beta_0$ equal to 0, but results are valid irrespective of $\beta_0$ value. Under this condition, all baseline expectations are aligned ($\mathbb{E}(p) = \Phi(\mathbb{E}(\eta)) = \mathbb{P}(\eta > 0) = 0.5 $). Example below estimates Shapley value for $x_1 = x_2 = c > 0$ and illustrates disagreement on the most important feature for different outcomes even when they agree on sign and there is no baseline difference.

\textbf{Shapley value for $\eta$}
\begin{align*}
    &\phi^\eta_1 = \phi^\eta_2 = c > 0\\
    &\phi^\eta_1 - \phi^\eta_2 = 0
\end{align*}
\textbf{Shapley value for Probability}
\begin{align*}
    \phi^p_1 &= \dfrac{1}{2}\left\{\left[\Phi\left(\dfrac{c}{\sqrt{\lambda +  \sigma^2_2}}\right) - 0.5\right] + \left[\Phi\left(\dfrac{2c}{\lambda}\right) - \Phi\left(\dfrac{c}{\sqrt{\lambda + \sigma^2_1}}\right)\right]\right\} > 0\\
    \phi^p_2 &= \dfrac{1}{2}\left\{\left[\Phi\left(\dfrac{c}{\sqrt{\lambda +  \sigma^2_1}}\right) - 0.5\right] + \left[\Phi\left(\dfrac{2c}{\lambda}\right) - \Phi\left(\dfrac{c}{\sqrt{\lambda + \sigma^2_2}}\right)\right]\right\} > 0\\
    \phi^p_1 - \phi^p_2 &= \Phi\left(\dfrac{c}{\sqrt{\lambda +  \sigma^2_2}}\right) - \Phi\left(\dfrac{c}{\sqrt{\lambda +  \sigma^2_1}}\right) > 0
\end{align*}
\textbf{Shapley value for Binary Decision}
\begin{align*}
    \phi^d_1 &= \dfrac{1}{2}\left\{\left[\Phi\left(\dfrac{c }{\sigma_2}\right) - 0.5\right] + \left[1 - \Phi\left(\dfrac{c}{\sigma_1}\right)\right]\right\} > 0\\
    \phi^d_2 &= \dfrac{1}{2}\left\{\left[\Phi\left(\dfrac{c}{\sigma_1}\right) - 0.5\right] + \left[1 - \Phi\left(\dfrac{c}{\sigma_2}\right)\right]\right\} > 0\\
    \phi^d_1 - \phi^d_2 &= \Phi\left(\dfrac{c}{\sigma_2}\right) - \Phi\left(\dfrac{c}{\sigma_1}\right) > 0
\end{align*}

Above expressions demonstrate that Shapley values of both features for $\eta$  are equal. But Shapley value for probability and Binary decision assign high value to feature with high variance (feature 1) even when all other parameters are same. This difference in Shapley value for features increases with increase in difference in their variance and the point $c$. Since the difference between features Shapley value is continuous and monotonic in feature value for all three outcomes, there exists a point at which there is disagreement between the most important feature of $\eta$ compared to probability/binary decision. Later we highlight the region of disagreement on the most important feature for other outcomes.

It seems counter intuitive that the Shapley value for probability/binary decision  for both features could be different even when their mean and regression coefficients are the same with $x_1 = x_2 = c$.  Here variance of the features is playing a crucial role. We can understand this with an example, consider a credit application model with regression intercept as 0, regression slope coefficient as 1, and features mean to be 0. Let the bank accept an application if the predicted probability is more than 0.5 equivalently 0 in $\eta$ and take $c=0.1$. Further assume, $\sigma_1$ is high but $\sigma_2$ is low such that feature 2 can not affect $\eta$ by more than $0.1$ with 95\% confidence. Although, both $x_1$ and $x_2$ are 0.1 but they are not equally important for the decision. We explain this by taking a hypothetical scenario where we know only one value. If we know $x_1 = 0.1$, then we will be almost certain that the application will get accepted as it is highly unlikely that $x_2$ reduces the probability below the acceptance threshold. But in the opposite case, where we know $x_2 = 0.1$, we are not confident that the application will get accepted because high variance in feature 1 may potentially decrease the probability value below acceptance threshold. This example signifies the role of variance in the feature importance.



To identify the region of disagreement on the most important feature, we start with estimating the curve for which absolute Shapley values  for both features  are equal. There are two cases in which both features can have equal Shapley values, first when Shapley values for both features have same sign ($\phi_1 = \phi_2$) and second, where Shapley values for both features have opposite sign  ($\phi_1 + \phi_2 = 0$). Below equations represent, equal importance curves for both cases (same sign and opposite sign) are linear in features value. Solution of $\phi_1 = \phi_2$ for different outcomes intersect at $(-\beta_0, -\beta_0)$, thus difference in equal importance curves with the same sign arises from the slope which does not depends on $\beta_0$. Solution of $\phi_1 + \phi_2 = 0$ for different outcomes are parallel to each other implying that the difference in equal importance curves with opposite sign arises from the intercept and when $\beta_0$ equals to 0 the lines for all outcomes are the same.
\begin{itemize}
    \item For $\eta$:
    \begin{itemize}
        \item same sign: $x_2 = x_1$
        \item opposite sign: $x_1 + x_2 = 0$
    \end{itemize}
    \item For probability: 
    \begin{itemize}
        \item same sign: $x_2 = \beta_0\left(\sqrt{\dfrac{\lambda + \sigma_1^2}{\lambda + \sigma_2^2}} - 1\right) + x_1\sqrt{\dfrac{\lambda + \sigma_1^2}{\lambda + \sigma_2^2}}$
        \item opposite sign: $x_1 + x_2 = -\beta_0 + \beta_0\sqrt{\dfrac{\lambda}{\lambda + \sigma_1^2 + \sigma_2^2}}$
    \end{itemize}
    \item For Binary Decision:
    \begin{itemize}
        \item same sign: $x_2 = \beta_0\left(\dfrac{\sigma_1}{\sigma_2} - 1\right) + x_1\dfrac{\sigma_1}{\sigma_2}$
        \item opposite sign\footnote{For binary decision outcome, sum of Shapley value is never equal to 0, but this sum has different sign for points above and below line $x_1 + x_2 = -\beta_0$. Thus we consider the $x_1 + x_2 = -\beta_0$ line as a solution for  $\phi_1 + \phi_2 = 0$. Note this simplification does not affect the disagreement results.}: $x_1 + x_2 = -\beta_0$
    \end{itemize}
    
\end{itemize}

\begin{figure}[H]
    \centering
    \caption{Equal Shapley Importance ($\beta_0 = 0, \sigma_1 = 2, \sigma_2 = 1$)}
    \includegraphics[scale=0.45, trim={0 0 0 1.3cm}, clip]{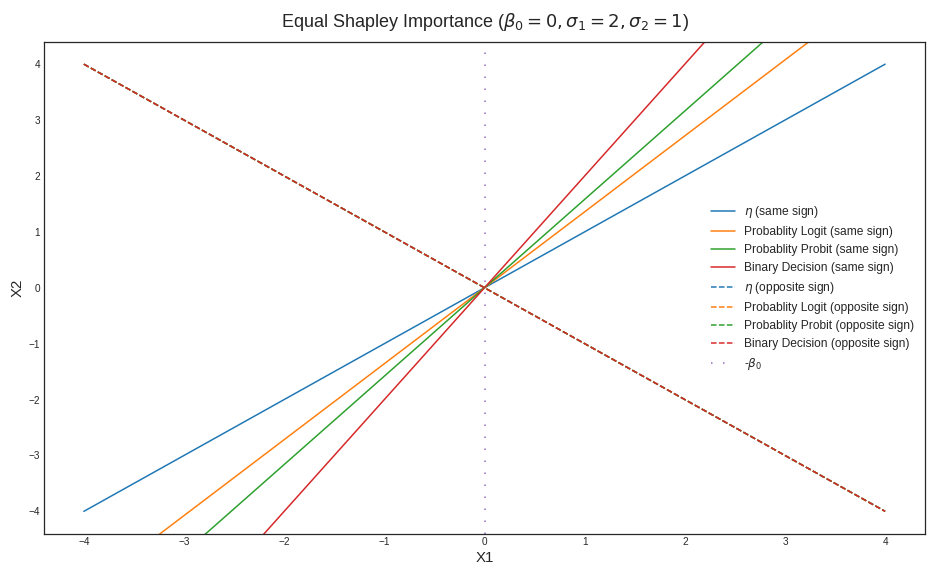}
    \label{fig:eq_shapley_boundary_beta0_0}
\end{figure}

\begin{figure}[H]
    \centering
    \caption{Equal Shapley Importance ($\beta_0 = 1, \sigma_1 = 2, \sigma_2 = 1$)}
    \includegraphics[scale=0.45, trim={0 0 0 1.3cm}, clip]{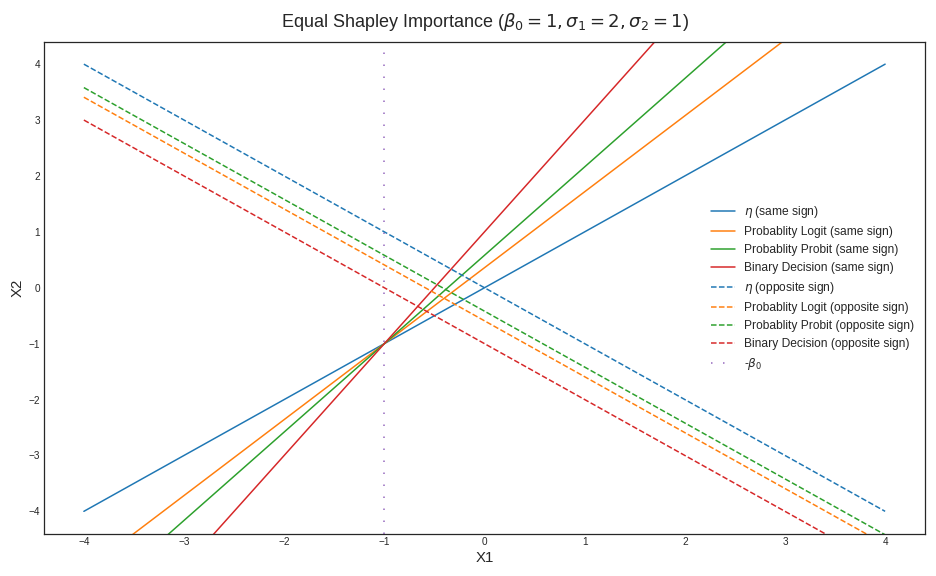} 
    \label{fig:eq_shapley_boundary_beta0_1}
\end{figure}

In Figure \ref{fig:eq_shapley_boundary_beta0_0} and \ref{fig:eq_shapley_boundary_beta0_1}, we indicate same sign equal importance curve with solid line and opposite sign equal importance curve with dashed line. For $\beta_0 = 0$ equal importance curves with opposite signs are exactly the same for all outcomes, the dashed lines overlap each other. Vertically bounded area between solid and dashed line (below solid line and above dashed line or above solid line and below dashed line) for particular outcome, is a region where feature 1 is more important than  feature 2\footnote{This holds because, Shapley value for both feature is 0 at the intersection of dashed and solid line. The partial derivative of feature $i$ Shapley value w.r.t. $X_i$ is positive and greater than or equal to  partial derivative of feature $i$ Shapley value w.r.t. $X_j$ for any feature $j$.}. Since the slope of equal Shapley explanation lines are different for different outcomes, there exists a region with disagreement on the most important feature. For example, a region bounded between red and blue solid line indicates that there is disagreement on the most important feature between $\eta$ and binary decision. For $\beta_0 = 1$, the dashed lines are parallel to each other as shown in Figure \ref{fig:eq_shapley_boundary_beta0_1}. Visually, a region of disagreement increases, as distance between dashed lines increases.


\section{Disagreements with Asymptotic Variance}\label{sec:diagreements_variance}
In the last section, we have discussed disagreements in Shapley values of different outcomes and how they depend on the value of $\beta_0$ and variance of variables. When  $\beta_0 = 0$, few disagreements disappear or weaken. In this section we will focus on how disagreements will behave when variance is high or low. When overall variance is low, most of the observations lie in a small region implying that the probability curve can be approximated by a linear line in that region. When overall volatility is high, there will be few observations around 0 (where derivative is high) making the binary decision curve a better candidate for probability curve approximation. Hence we expect, for low variance  Shapley value for probability and $\eta$  to be similar and for high variance Shapley value for probability and binary decision outcomes to be similar.

Although, we have normalized the mean of the features to 0 and their regression coefficient to 1 but in this section, we write an equation using the mean and regression coefficient to make it explicit. For probability outcome we will use logit regression but the same results hold for probit model.

\begin{figure}[H]
    \centering
    \caption{Baseline expectation logit model ($\mathbb{E}(\eta) = \beta_0 + \beta_1\mu_1 + \beta_2\mu_2 = 1$)}
    \includegraphics[scale=0.45, trim={0 0 0 1.3cm}, clip]{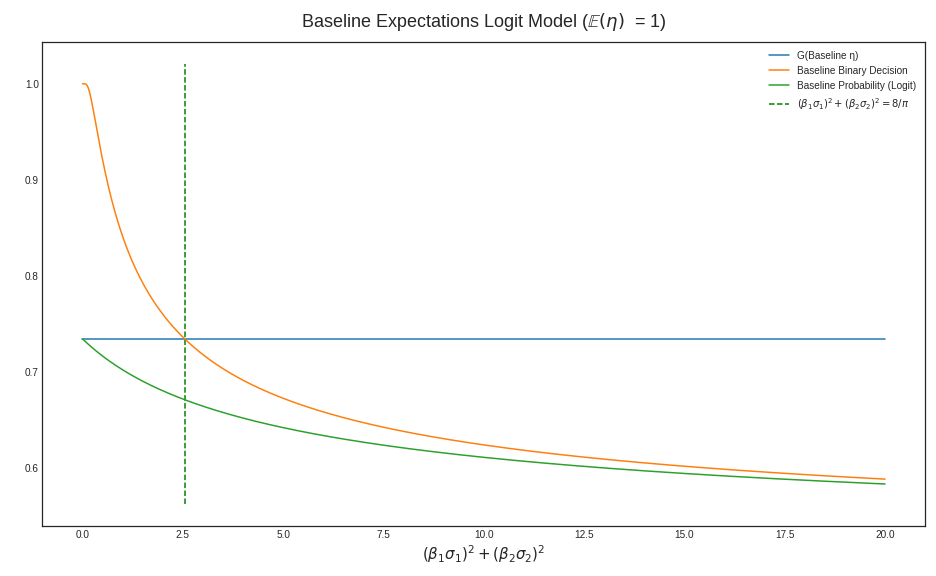}
    \label{fig:baseline_expectations_logit}
\end{figure}

Figure \ref{fig:baseline_expectations_logit} illustrates the baseline expectation for different outcomes when mean for $\eta$ is positive ($\mathbb{E}(\eta) = \beta_0 + \beta_1\mu_1 + \beta_2\mu_2 > 0$). For low variance\footnote{Overall model variance can be measured by the variance of $\eta$ which is $\mathbb{V}(\eta) = (\beta_1\sigma_1)^2 + (\beta_2\sigma_2)^2$}, the baseline expectation for probability outcome is close to $\eta$ and for high variance the baseline expectation for probability outcome is close to binary decision.

In order to measure the scale of disagreement on sign and most important feature, we have simulated 1 million sample points for different expected value and variance of log-odds. Table \ref{tab:sign_disagreement} and \ref{tab:important_feature_disagreement} describe the percentage of sample with sign disagreement and the percentage of sample with disagreement on most important feature respectively\footnote{When expected value  of $\eta$ is high with low variance, the binary decision outcome is always positive, i.e., model outcome for binary decision is constant implying 0 Shapley value for all features. Hence, the table is left blank for binary decision with parameter $\beta_1 = 1, \beta_1\sigma_1=0.02, \beta_2\sigma_2 = 0.01$}\textsuperscript{,}\footnote{For table 7 and 8 we take $\beta_1\sigma_1  = 2\beta_2\sigma_2$. Result can slightly differ for different parameter but conclusions are general.}. In the case of low variance ($\beta_1\sigma_1 = 0.02$ and $\beta_2\sigma_2 = 0.01$), there is almost no disagreement between log-odds and probability outcome and in the case of high variance ($\beta_1\sigma_1 = 200$ and $\beta_2\sigma_2 = 100$) there is almost no disagreement between probability and binary decision outcome. Both of these tables also highlight that disagreement percentage increases with increase in expected value of $\eta$ (leaving very high variance case). More details on disagreement for high/low variance case is available in Appendix \ref{sec:appendix_disagreements}.

\begin{table}[H]
\caption{Disagreement on sign}
\label{tab:sign_disagreement}
\centering
\begin{tabular}{|c|c|c|r|r|r|}
\hline
\multirow{4}{*}{$\bm{\mathbb{E}(\eta)}$} & \multirow{4}{*}{$\bm{\beta_1\sigma_1}$} & \multirow{4}{*}{$\bm{\beta_2\sigma_2}$} & \multicolumn{3}{c|}{\textbf{Sign Disagreement}} \\ \cline{4-6} 
 & & & \multicolumn{1}{c|}{\textbf{\begin{tabular}[c]{@{}c@{}}log-odds \\ vs \\ probability\end{tabular}}} & \multicolumn{1}{c|}{\textbf{\begin{tabular}[c]{@{}c@{}}probability \\ vs \\ binary-decision\end{tabular}}} & \multicolumn{1}{c|}{\textbf{\begin{tabular}[c]{@{}c@{}}log-odds \\ vs \\ binary-decision\end{tabular}}} \\ \hline
\multirow{3}{*}{0} & 0.02 & 0.01 & \cellcolor{blue!25}0.00\% & 21.63\% & 21.63\% \\ \cline{2-6} 
 & 2 & 1 & 6.45\% & 17.51\% & 21.63\% \\ \cline{2-6} 
 & 200 & 100 & 21.34\% &\cellcolor{blue!25} 0.30\% & 21.63\% \\ \hline
\multirow{3}{*}{1} & 0.02 & 0.01 & \cellcolor{blue!25}0.23\% & \multicolumn{1}{l|}{} & \multicolumn{1}{l|}{} \\ \cline{2-6} 
 & 2 & 1 & 10.71\% & 16.81\% & 23.94\% \\ \cline{2-6} 
 & 200 & 100 & 21.33\% &\cellcolor{blue!25} 0.30\% & 21.63\% \\ \hline
\end{tabular}
\end{table}

\begin{table}[H]
\caption{Disagreement on Most Important Feature}
\label{tab:important_feature_disagreement}
\centering
\begin{tabular}{|c|c|c|r|r|r|}
\hline
\multirow{4}{*}{$\bm{\mathbb{E}(\eta)}$} & \multirow{4}{*}{$\bm{\beta_1\sigma_1}$} & \multirow{4}{*}{$\bm{\beta_2\sigma_2}$} & \multicolumn{3}{c|}{\textbf{Important Feature Disagreement}} \\ \cline{4-6} 
 & & & \multicolumn{1}{c|}{\textbf{\begin{tabular}[c]{@{}c@{}}log-odds \\ vs \\ probability\end{tabular}}} & \multicolumn{1}{c|}{\textbf{\begin{tabular}[c]{@{}c@{}}probability \\ vs \\ binary-decision\end{tabular}}} & \multicolumn{1}{c|}{\textbf{\begin{tabular}[c]{@{}c@{}}log-odds \\ vs \\ binary-decision\end{tabular}}} \\ \hline
\multirow{3}{*}{0} & 0.02 & 0.01 &\cellcolor{blue!25} 0.00\% & 6.94\% & 6.95\% \\ \cline{2-6} 
 & 2 & 1 & 3.53\% & 3.42\% & 6.95\% \\ \cline{2-6} 
 & 200 & 100 & 6.95\% & \cellcolor{blue!25}0.00\% & 6.95\% \\ \hline
\multirow{3}{*}{1} & 0.02 & 0.01 &\cellcolor{blue!25} 0.11\% & \multicolumn{1}{l|}{} & \multicolumn{1}{l|}{} \\ \cline{2-6} 
 & 2 & 1 & 5.53\% & 5.24\% & 10.77\% \\ \cline{2-6} 
 & 200 & 100 & 6.94\% &\cellcolor{blue!25} 0.00\% & 6.94\% \\ \hline
\end{tabular}
\end{table}

\section{Global Feature Importance} \label{sec:global_feature_importance}
Global feature importance for a model gives a number estimate for the importance of a feature at global level. It is commonly used to compare the global relevance of a feature and to understand which features are more relevant compared to others. Global feature importance is calculated  by taking the sum of absolute Shapley value of the feature over all sample, i.e.,
\[\mathcal{I}_i = \sum_{j = 1}^{n_{sample}} \left|\phi_{j, i}\right|\]
where, $\mathcal{I}_i$ denotes the global importance of feature $i$ and $\phi_{j, i}$ denotes the Shapley value of feature $i$ for sample $j$. Global importance of feature $i$ for $\eta$ outcome (denoted  by $\mathcal{I}_i^\eta$) is given below\footnote{F. C. Leone. \cite{leone1961folded} has shown for $Z \sim \mathcal{N}(0, \ \sigma^2), \mathbb{E}\ |Z| = \sigma\sqrt{\dfrac{2}{\pi}}$. This imply $\sum_j |z_j| \approx \sigma\times n_{sample}\sqrt{\dfrac{2}{\pi}}$}. Since global importance for the feature is scalar multiple of $\beta_i\sigma_i$, relative global importance of the features for $\eta$ rely on their relative $\beta_i\sigma_i$.
\begin{align*}
    \mathcal{I}_i^\eta = \sum_{j = 1}^{n_{sample}}|\beta_i(x_i^j - \mu_i)| = \beta_i\sigma_i\times n_{sample}\sqrt{\dfrac{2}{\pi}}
\end{align*}

For probability and binary decision outcomes, global feature importance does not have any closed form solution. Thus, we have simulated 1 million samples with different expected values and variance of log-odds. Table \ref{tab:global_feature_importance} illustrates the relative global importance of feature 1 for all outcomes and excess relative global feature importance of feature 1 for probability and binary decision outcomes compared to log-odds. Relative global importance of feature 1 for $\eta$ equals $\nicefrac{\beta_1\sigma_1}{\beta_2\sigma_2}$  which is taken as 2 or 5. This table is displaying that probability and binary decision outcome assign higher relative importance to feature 1 compared to the log-odds outcome. Probability outcome is excessing relative global importance of feature 1 by 28\% compared to log-odd outcome when $\beta_1\sigma_1 = 5$ and $\beta_2\sigma_2 = 1$. In case of low variance, global feature importance of log-odds and probability outcomes are equal and for high variance, global feature importance of probability and binary decision outcomes are equal.

\begin{table}[H]
\caption{Global Feature Importance}
\label{tab:global_feature_importance}
\centering
\begin{changemargin}{-7mm}{-7mm}
\begin{tabular}{|c|c|c|r|r|r|r|r|}
\hline
\multirow{2}{*}{\bm{$\mathbb{E}(\eta)$}}& \multirow{2}{*}{\bm{$\beta_1\sigma_1$}}& \multirow{2}{*}{\bm{$\beta_2\sigma_2$}}& \multicolumn{3}{c|}{\textbf{Relative Feature Importance}} & \multicolumn{2}{c|}{\textbf{Excess Relative Importance}} \\ \cline{4-8} 
& & & \multicolumn{1}{c|}{\textbf{log odds}} & \multicolumn{1}{c|}{\textbf{probability}}& \multicolumn{1}{c|}{\textbf{binary decision}} & \multicolumn{1}{c|}{\textbf{probability}}&\multicolumn{1}{c|}{\textbf{binary decision}} \\ \hline
\multirow{3}{*}{0} & 0.02 & 0.01 & 2.00 & 2.00 & 2.28 & 0.0\% & 13.8\% \\ \cline{2-8} 
& 2 & 1 & 2.00 & 2.19 & 2.28 & 9.5\% & 13.8\% \\ \cline{2-8} 
& 200 & 100 & 2.00 & 2.28 & 2.28 & \cellcolor{red!25} 14.0\% & 13.8\% \\ \hline
\multirow{3}{*}{1} & 0.02 & 0.01 & 2.00 & 2.00 & \multicolumn{1}{l|}{} & 0.0\% & \\ \cline{2-8} 
& 2 & 1 & 2.00 & 2.18 & 2.24 & 9.0\% & 12.1\% \\ \cline{2-8} 
& 200 & 100 & 2.00 & 2.28 & 2.28 & 14.0\% & 13.8\% \\ \hline
\multirow{3}{*}{0} & 0.05 & 0.01 & 5.00 & 5.00 & 6.26 & 0.0\% & 25.3\% \\ \cline{2-8} 
& 5 & 1 & 5.00 & 6.41 & 6.26 & \cellcolor{red!25}28.3\% & 25.3\% \\ \cline{2-8} 
& 500 & 100 & 5.00 & 6.28 & 6.26 & 25.7\% & 25.3\% \\ \hline
\multirow{3}{*}{1} & 0.05 & 0.01 & 5.00 & 5.00 & \multicolumn{1}{l|}{} & 0.0\% & \multicolumn{1}{l|}{} \\ \cline{2-8} 
& 5 & 1 & 5.00 & 6.39 & 6.23 & 27.9\% & 24.5\% \\ \cline{2-8} 
& 500 & 100 & 5.00 & 6.29 & 6.26 &25.7\% & 25.3\% \\ \hline
\end{tabular}
\end{changemargin}
\end{table}

\section{Conclusion}\label{sec:conclusion}
Shapley value is a  method for explaining the contribution of features in prediction with a game theoretical foundation with certain desirable properties  of fairness. To understand model prediction, it is essential to understand the contribution of its features and Shapley values assign that contribution value to features. We have already discussed what are the factors that influence the Shapley value of a linear probability model for different outcomes such as probability, log-odds, binary decision. 

Our principal findings include Shapley value for probability and binary decision outcomes depend on overall variance and other features value unlike log-odds where Shapley explanation is a function of its mean, regression coefficient and feature value. Moreover, Shapley value for binary decision is discontinuous. There are disagreements in Shapley value for different outcomes, such as baseline expectation for Probability/log-odds/binary decision outcomes are different implying different reference points, which make these values incomparable in terms of inter-model outcome. Sign of Shapley value for the same feature can be different for different outcomes because relevance of variance is different for different outcomes. Even most important features can vary for different outcomes suggesting that feature A can hold top importance for Probability outcome but might not be on top for decision making (accept or reject) outcome. These disagreements over Shapley values between probability and log odds outcomes dissolve if overall variance is low. When overall variance is high then there are minimal disagreements in Shapley values for Probability and binary decision outcomes. Global feature importance for probability and binary decision outcomes is more influenced with variance compared to log-odds outcome. 

In credit risk modeling the same model can be used to predict different outcomes. Given, there is no unique Shapley explanation for a model as they vary for different outcomes with some disagreements. We should estimate the Shapley explanation according to the usages of the model. For example, if the model is used for accepting or rejecting the loan application then Shapley explanation for binary decision is more suitable. If we are estimating probability of default then Shapley explanation for probability outcome is more appropriate. These conclusions are not only limited to linear probability model but are applicable to broader class of machine learning models.

\pagebreak

\printbibliography
\pagebreak
\begin{appendix}
\section{Derivation of Shapley value}\label{sec:appendix_shapley_value}
{\Large \textbf{Shapley value for} $\eta$:}\\
Value function for $\eta$ is:
\begin{align*}
    v^\eta(\{\}) &= \mathbb{E}[\eta] =  \beta_0 + \beta_1\mu_1 + \beta_2\mu_2\\
    v^\eta(\{X_1\}) &= \mathbb{E}(\eta \ | X_1 = x_1) = \beta_0 + \beta_1x_1 + \beta_2\mu_2\\
    v^\eta(\{X_2\}) &= \mathbb{E}(\eta \ | X_2 = x_2) = \beta_0 + \beta_1\mu_1 + \beta_2x_2\\
    v^\eta(\{X_1, X_2\}) &= \mathbb{E}(\eta\ | X_1 =x_1, X_2 = x_2) = \beta_0 + \beta_1x_1 + \beta_2x_2
\end{align*}

Now we apply Shapley value formula to estimate the value for each feature
\begin{align*}
    \phi^\eta_0 &= v^\eta(\{\}) = \beta_0 + \beta_1\mu_1 + \beta_2\mu_2\\
    \phi^\eta_1 &= \dfrac{1}{2}\left\{\left[v^\eta(\{X_1\}) - v^\eta(\{\})\right] + \left[v^\eta(\{X_1, X_2\}) - v^\eta(\{X_2\})\right]\right\}\\
    \phi^\eta_2 &= \dfrac{1}{2}\left\{\left[v^\eta(\{X_2\}) - v^\eta(\{\})\right] + \left[v^\eta(\{X_1, X_2\}) - v^\eta(\{X_1\})\right]\right\}\\[5mm]
    \implies \phi^\eta_0 &= \beta_0 + \beta_1\mu_1 + \beta_2\mu_2\\
    \phi^\eta_1 &= \beta_1(x_1 - \mu_1)\\
    \phi^\eta_2 &= \beta_2(x_2 - \mu_2)
\end{align*}

{\Large \textbf{Shapley value for Binary Decision} ($\mathbbm{1}\left(\eta \ge \eta^*\right)$ or $\mathbbm{1}\left(p \ge p^*\right)$):}\\
Value function for Binary Decision outcome equals to
\begin{align*}
    v^d(\{\}) &= \mathbb{E}[\mathbbm{1}\left(\eta \ge \eta^*\right)] =  Pr(X\beta \ge \eta^*) = 1 - \Phi\left(\dfrac{ \eta^* - \beta_0  - \beta_1\mu_1 - \beta_2\mu_2}{\sqrt{\beta^2_1\sigma^2_1 + \beta^2_2\sigma^2_2}}\right)\\
    \implies v^d(\{\}) &= \mathbb{P}(X\beta \ge \eta^*) = \Phi\left(\dfrac{\beta_0 + \beta_1\mu_1 + \beta_2\mu_2 - \eta^*}{\sqrt{\beta^2_1\sigma^2_1 + \beta^2_2\sigma^2_2}}\right)\\
    v^d(\{X_1\}) &= \mathbb{P}(X\beta \ge \eta^* \ | X_1 = x_1) = \Phi\left(\dfrac{\beta_0 + \beta_1x_1 + \beta_2\mu_2 - \eta^*}{\beta_2\sigma_2}\right)\\
    v^d(\{X_2\}) &= \mathbb{P}(X\beta \ge \eta^* \ | X_2 = x_2) = \Phi\left(\dfrac{\beta_0 + \beta_1\mu_1 + \beta_2x_2 - \eta^*}{\beta_1\sigma_1}\right)\\
    v^d(\{X_1, X_2\}) &= \mathbb{P}(X\beta \ge \eta^* \ | X_1 =x_1, X_2 = x_2) = \mathbbm{1}\left(\beta_0 + \beta_1x_1 + \beta_2x_2 \ge \eta^*\right)
\end{align*}

Now we apply the Shapley value formula to estimate the value for each feature
\begin{align*}
    \phi^d_0 &= \Phi\left(\dfrac{\beta_0 + \beta_1\mu_1 + \beta_2\mu_2 - \eta^*}{\sqrt{\beta^2_1\sigma^2_1 + \beta^2_2\sigma^2_2}}\right)\\
    \phi^d_1 &= \dfrac{1}{2}\left\{\left[\Phi\left(\dfrac{\beta_0 + \beta_1x_1  + \beta_2\mu_2 - \eta^*}{\beta_2\sigma_2}\right) - \Phi\left(\dfrac{\beta_0 + \beta_1\mu_1 + \beta_2\mu_2 - \eta^*}{\sqrt{\beta^2_1\sigma^2_1 + \beta^2_2\sigma^2_2}}\right)\right] \right. \\
    &\left. \quad \quad \quad + \left[\mathbbm{1}\left(\beta_0 + \beta_1x_1 + \beta_2x_2 \ge \eta^*\right) - \Phi\left(\dfrac{\beta_0 + \beta_1\mu_1 + \beta_2x_2 - \eta^*}{\beta_1\sigma_1}\right)\right]\right\} \\[3mm]
    \phi^d_2 &= \dfrac{1}{2}\left\{\left[\Phi\left(\dfrac{\beta_0 + \beta_1\mu_1  + \beta_2x_2 - \eta^*}{\beta_1\sigma_1}\right) - \Phi\left(\dfrac{\beta_0 + \beta_1\mu_1 + \beta_2\mu_2 - \eta^*}{\sqrt{\beta^2_1\sigma^2_1 + \beta^2_2\sigma^2_2}}\right)\right] \right. \\
    &\left. \quad \quad \quad + \left[\mathbbm{1}\left(\beta_0 + \beta_1x_1 + \beta_2x_2 \ge \eta^*\right) - \Phi\left(\dfrac{\beta_0 + \beta_1x_1 + \beta_2\mu_2 - \eta^*}{\beta_2\sigma_2}\right)\right]\right\}\\
\end{align*}

{\Large \textbf{Shapley value for Probability} ($p$):}\\
In order to estimate the value function for the logit/probit model, we need to simplify $E[G(Z)]$, where $Z$ is distributed normally. In the case of probit model ($G$ is standard normal),  a simplified expression for expectation is described below\cite{ellison1964two} .
\[\mathbb{E}[\Phi(Z)] = \Phi\left(\dfrac{\mu}{\sqrt{1 + \sigma^2}}\right) \quad \quad \text{where, } Z \sim \mathcal{N}(\mu, \sigma^2)\]
 This expectation does not have a closed form solution in the case of logit regression ($G$ is logistic distribution). Hence we use standard normal function approximation (described below) for Logistic function also known as sigmoid function denoted  by $S$\cite{tocher}\cite{dombi2018approximations}. 
\[S(x) \approx \Phi\left(\dfrac{x}{\sqrt{\nicefrac{8}{\pi}}}\right) \]
Using above approximation, we can simplify $E[G(Z)]$ as
\[\mathbb{E}[G(Z)] = \Phi\left(\dfrac{\mu}{\sqrt{\lambda + \sigma^2}}\right) \quad \quad \text{where, } Z \sim \mathcal{N}(\mu, \sigma^2)\]
where, $\lambda$ equals to 1 in case of Probit model ($G = \Phi$) and $\lambda = \nicefrac{8}{\pi} \approx 2.5465$ for the Logit model ($G = S$). \\

Value function for Shapley explanation for probability outcome is 
\begin{align*}
    v^p(\{\}) &= \mathbb{E}[\Phi(X\beta)] = \Phi\left(\dfrac{\beta_0 + \beta_1\mu_1 + \beta_2\mu_2}{\sqrt{\lambda + \beta^2_1\sigma^2_1 + \beta^2_2\sigma^2_2}}\right)\\
     v^p(\{X_1\}) &= \mathbb{E}[\Phi(X\beta) | X_1 = x_1] = \Phi\left(\dfrac{\beta_0 + \beta_1x_1 + \beta_2\mu_2}{\sqrt{\lambda + \beta^2_2\sigma^2_2}}\right)\\
     v^p(\{X_2\}) &= \mathbb{E}[\Phi(X\beta) | X_2 = x_2]  = \Phi\left(\dfrac{\beta_0 + \beta_1\mu_1 + \beta_2x_2}{\sqrt{\lambda + \beta^2_1\sigma^2_1}}\right) \\
     v^p(\{X_1, X_2\}) &= \mathbb{E}[\Phi(X\beta) | X_1 = x_1, X_2 = x_2]  = \Phi\left(\dfrac{\beta_0 + \beta_1x_1 + \beta_2x_2}{\sqrt{\lambda}}\right)
\end{align*}

Now we apply Shapley value formula to estimate the value for each feature
\begin{align*}
    \phi^p_0 &= \Phi\left(\dfrac{\beta_0 + \beta_1\mu_1 + \beta_2\mu_2}{\sqrt{\lambda + \beta^2_1\sigma^2_1 + \beta^2_2\sigma^2_2}}\right)\\
    \phi^p_1 &= \dfrac{1}{2}\left\{\left[\Phi\left(\dfrac{\beta_0 + \beta_1x_1  + \beta_2\mu_2}{\sqrt{\lambda + \beta^2_2\sigma^2_2}}\right) - \Phi\left(\dfrac{\beta_0 + \beta_1\mu_1 + \beta_2\mu_2}{\sqrt{\lambda + \beta^2_1\sigma^2_1 + \beta^2_2\sigma^2_2}}\right)\right] \right. \\
    &\left. \quad \quad \quad + \left[\Phi\left(\dfrac{\beta_0 + \beta_1x_1 + \beta_2x_2}{\sqrt{\lambda}}\right) - \Phi\left(\dfrac{\beta_0 + \beta_1\mu_1 + \beta_2x_2}{\sqrt{\lambda + \beta^2_1\sigma^2_1}}\right)\right]\right\} \\[3mm]
    \phi^p_2 &= \dfrac{1}{2}\left\{\left[\Phi\left(\dfrac{\beta_0 + \beta_1\mu_1  + \beta_2x_2}{\sqrt{\lambda + \beta^2_1\sigma^2_1}}\right) - \Phi\left(\dfrac{\beta_0 + \beta_1\mu_1 + \beta_2\mu_2}{\sqrt{\lambda + \beta^2_1\sigma^2_1 + \beta^2_2\sigma^2_2}}\right)\right] \right. \\
    &\left. \quad \quad \quad + \left[\Phi\left(\dfrac{\beta_0 + \beta_1x_1 + \beta_2x_2}{\sqrt{\lambda}}\right) - \Phi\left(\dfrac{\beta_0 + \beta_1x_1 + \beta_2\mu_2}{\sqrt{\lambda + \beta^2_2\sigma^2_2}}\right)\right]\right\}
\end{align*}

\pagebreak
\section{Appendix for Disagreements with Asymptotic Variance} \label{sec:appendix_disagreements}
In this appendix, We have discussed disagreement on sign and most important feature in the case of high and low variance.  For this we have used simplified expression i.e., after substituting $\mu_i = 0$, and $\beta_i$ = 1 for $i \in {1, 2}$.

\begin{figure}[H]
    \centering
    \caption{Level Curve for zero Shapley value ($\beta_0$ = 0, $\sigma_1$ = 0.02, $\sigma_2$ = 0.01)}
    \includegraphics[scale=0.45, trim={0 0cm 0 1.3cm}, clip]{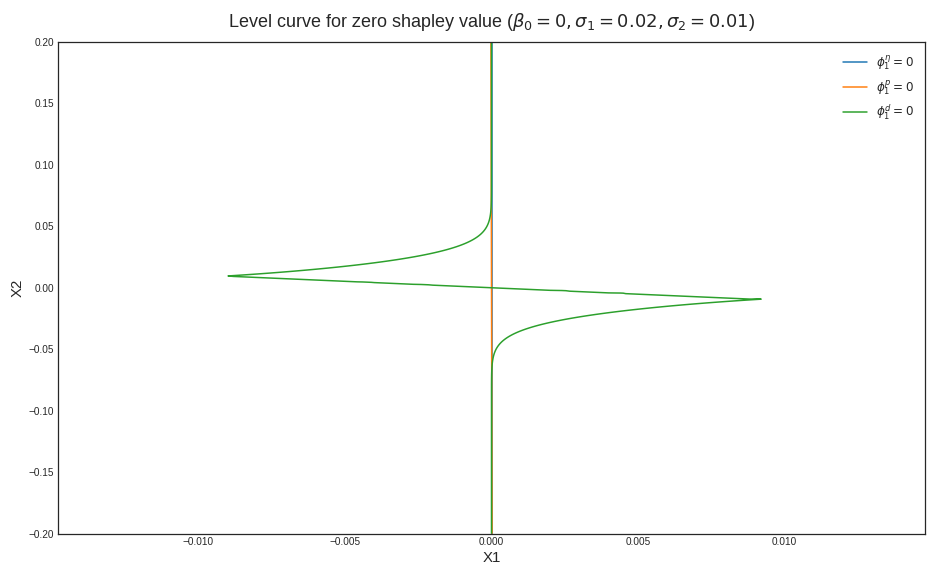}
    \label{fig:level_curve_zero_shap_low_var_beta0_0}
\end{figure}

\begin{figure}[H]
    \centering
    \caption{Level Curve for zero Shapley value ($\beta_0$ = 0.05, $\sigma_1$ = 0.02, $\sigma_2$ = 0.01)}
    \includegraphics[scale=0.45, trim={0 0.3cm 0 1.3cm}, clip]{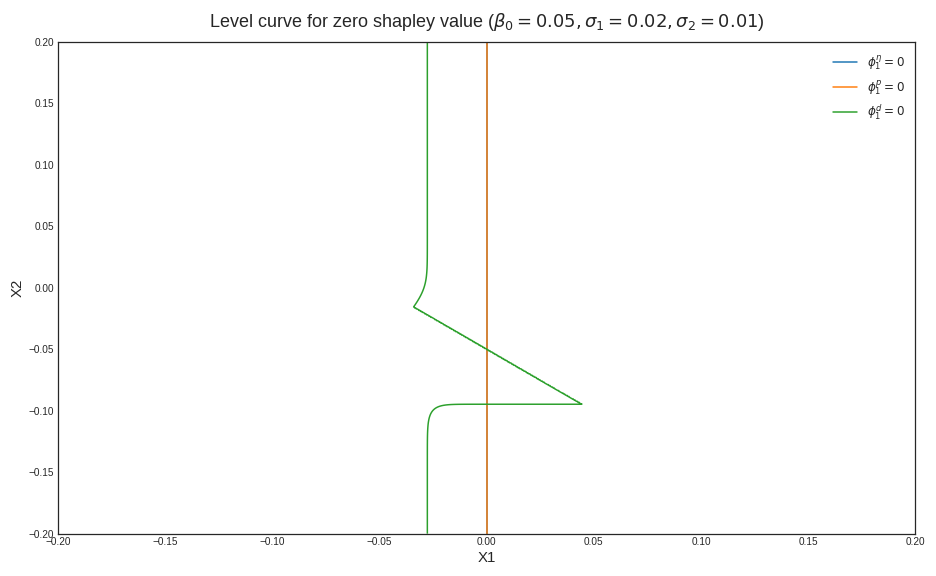}
   \label{fig:level_curve_zero_shap_low_var_beta0_pos}
\end{figure}

{\large\textbf{Disagreement in Sign}}\\
Figure \ref{fig:level_curve_zero_shap_low_var_beta0_0} and \ref{fig:level_curve_zero_shap_low_var_beta0_pos} illustrate the level curve for zero Shapley value for different outcome for low variance case ($\sigma_1$ = 0.02, $\sigma_2$ = 0.01) with $\beta_0$ equal to 0 and 0.05. Both figures depict that the level curve for $\eta$ and probability collide with each other.  

\begin{figure}[H]
    \centering
    \caption{Level Curve for zero Shapley value ($\beta_0$ = 0, $\sigma_1$ = 20, $\sigma_2$ = 10)}
    \includegraphics[scale=0.45, trim={0 0 0 1.3cm}, clip]{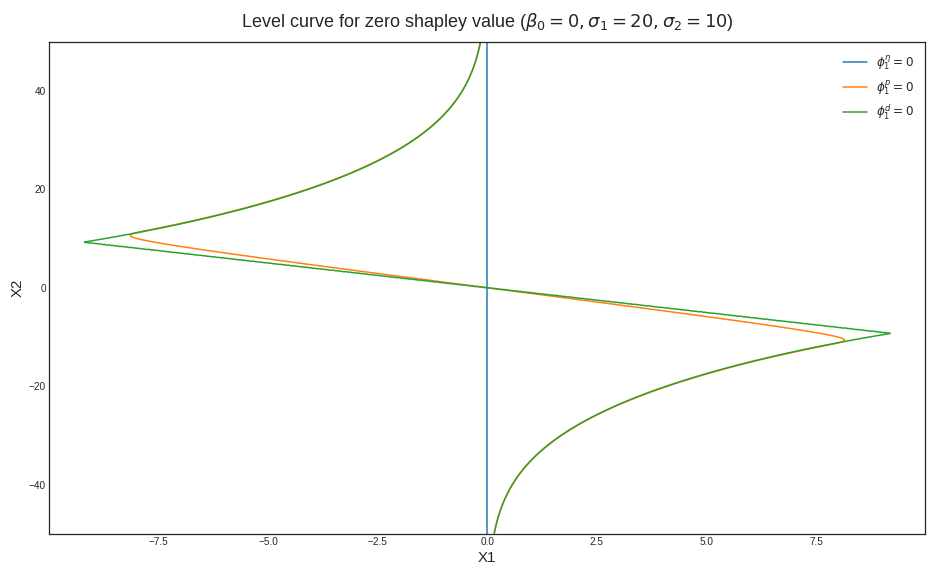}
    \label{fig:level_curve_zero_shap_high_var_beta0_0}
\end{figure}

\begin{figure}[H]
    \centering
    \caption{Level Curve for zero Shapley value ($\beta_0$ = 10, $\sigma_1$ = 20, $\sigma_2$ = 10)}
    \includegraphics[scale=0.45, trim={0 0 0 1.3cm}, clip]{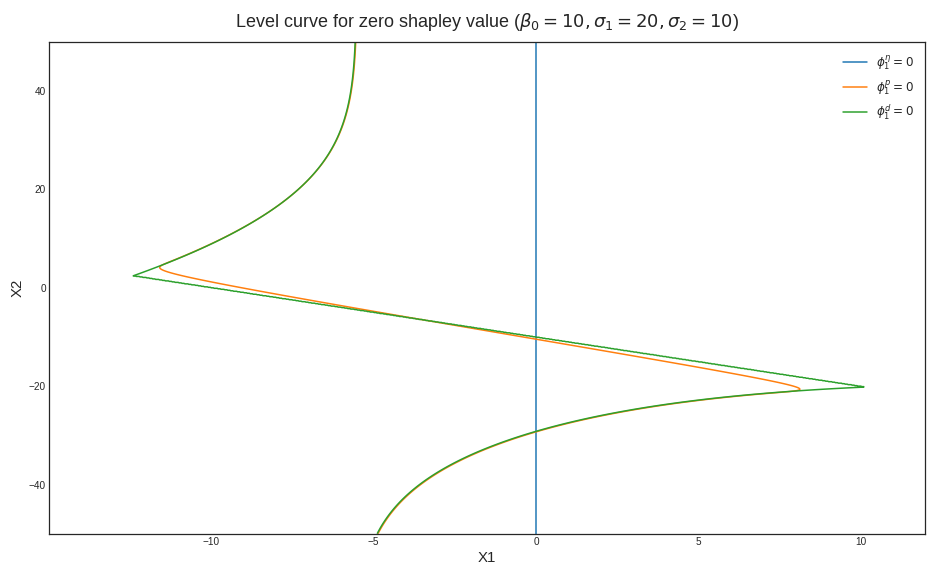}  
    \label{fig:level_curve_zero_shap_high_var_beta0_pos}
\end{figure}

Figure \ref{fig:level_curve_zero_shap_high_var_beta0_0} and \ref{fig:level_curve_zero_shap_high_var_beta0_pos} display the level curve for zero Shapley value for different outcome for high variance case ($\sigma_1$ = 20, $\sigma_2$ = 10) with $\beta_0$ equal to 0 and 10. We can see  that the level curve for probability and binary decision are almost same. 

\begin{figure}[H]
    \centering
    \caption{Relationship between variance and slope\\ for equal importance curve with same sign  ($\sigma_1 = 2\sigma_2 $)}
    \includegraphics[scale=0.45, trim={0 0 0 1.3cm}, clip]{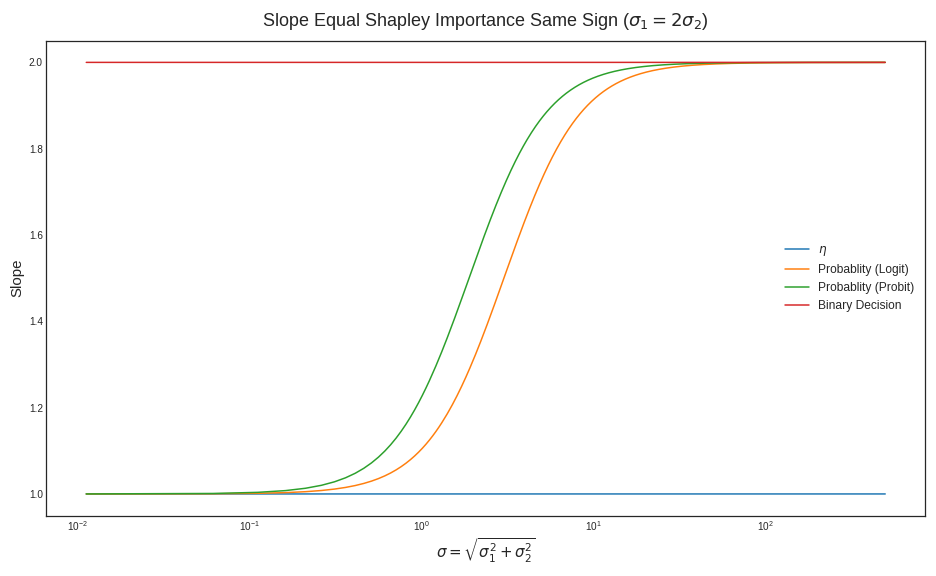}
    \label{fig:equal_importance_curve_same_sign_slope}
\end{figure}

\begin{figure}[H]
    \centering
    \caption{Relationship between variance and intercept \\ for equal importance curve with opposite sign ($\beta_0 = 1$)}
    \includegraphics[scale=0.45, trim={0 0 0 1.3cm}, clip]{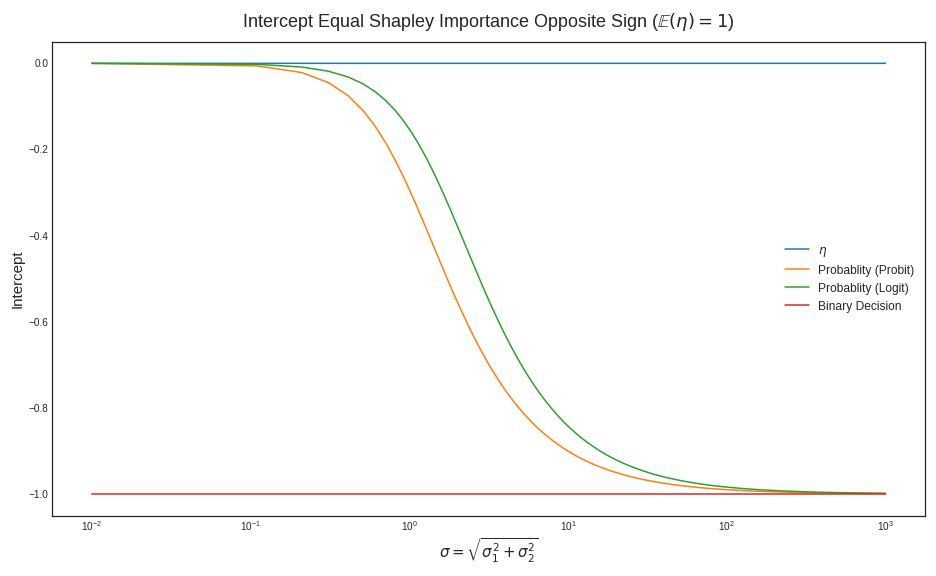}
    \label{fig:equal_importance_curve_opposite_sign_intercept}
\end{figure}

\vspace{4mm}
{\large\textbf{Disagreement in Most Important Feature}}\\
The disagreement on the equal importance with the same sign arises due to different slope coefficients and disagreement on the equal importance with the opposite sign arises due to different intercept value. Figure \ref{fig:equal_importance_curve_same_sign_slope} portrays a relationship between variance and slope for equal Shapley importance curve with same sign. Figure \ref{fig:equal_importance_curve_opposite_sign_intercept} represents a relationship between variance and intercept for equal Shapley importance curve with opposite sign. For low variance, probability and $\eta$ will have the same equal importance curve whereas when variance is high probability and binary decision outcomes will have matching equal importance curves.

\end{appendix}
\end{document}